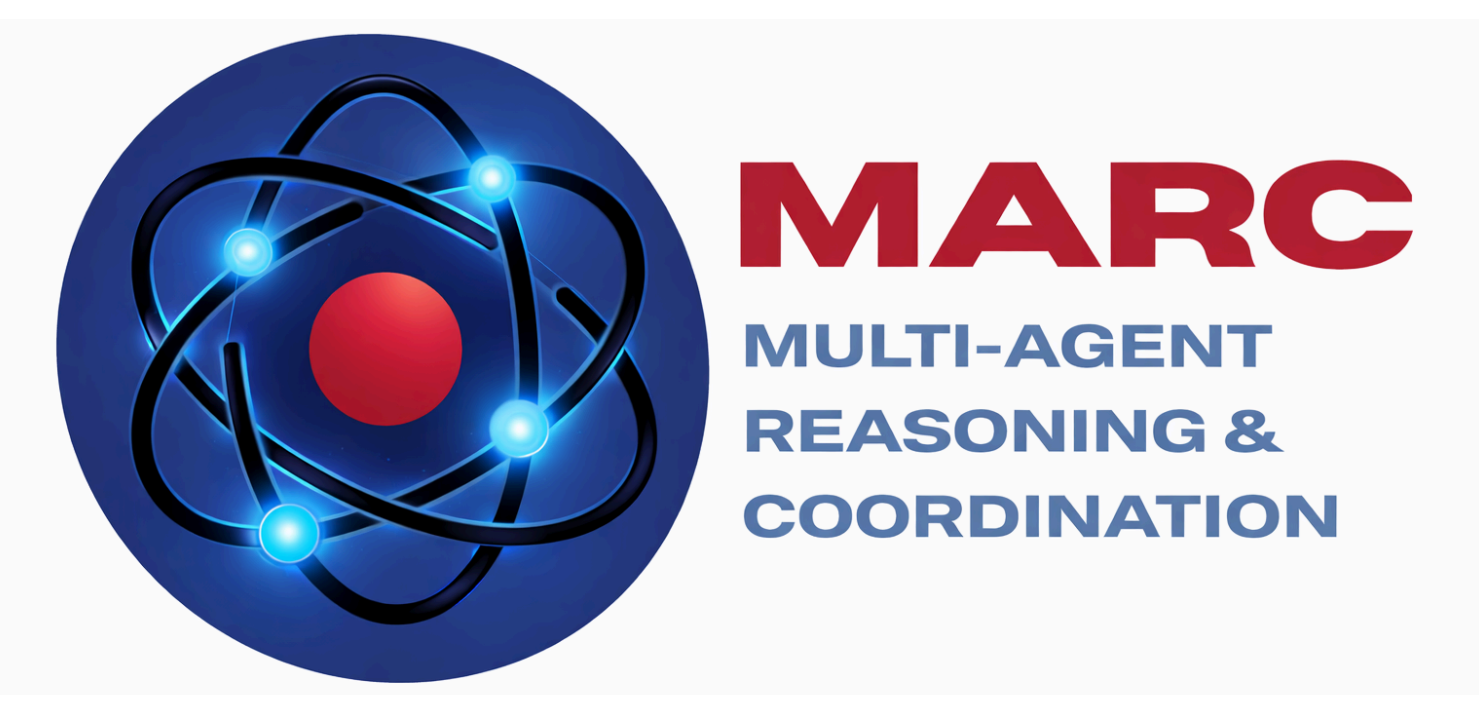

# MARC v1: An Open-Source Multi-Agent Framework for Clinical AI Reasoning and Coordination

Saisha Shetty[1+], Satvik Tripath[2+*], Austin Lin[3], Colin Zhao[3], Theodore Kim[3], Don Enwerem[4], Jacinta Arnold[5], Shahriar Faghani[2], Tessa S Cook[2]

1. College of Engineering, University of California, Davis
2. Perelman School of Medicine, University of Pennsylvania
3. School of Engineering and Applied Science, University of Pennsylvania
4. College of Computing and Informatics, Drexel University
5. UC Davis Graduate School of Management, Davis, CA

+ co-first authors
*satvik.tripathi@pennmedicine.upenn.edu

---

**Abstract**

We present Multi-Agent Reasoning and Coordination (MARC), an open-source framework that replaces monolithic LLM prompting with deterministic multi-agent orchestration for clinical reasoning. MARC coordinates role-specialized agents for extraction, reasoning, answer generation, and evaluation, with explicit context passing and traceable intermediate outputs, enabling stage-wise failure attribution. We additionally introduce a Decomposer module that generates task-specific agent prompts from a plain-language description, eliminating manual prompt engineering. The framework supports both API-based and local CPU-compatible deployments and is entirely configurable, without code modifications. MARC is designed to be model-agnostic, interpretable, and accessible to clinical domain experts without programming expertise. The full framework is available at https://github.com/Penn-RAIL/MARC-v1.

---

## 1. Introduction

Artificial intelligence and large language models (LLMs) have seen rapid adoption across medicine, demonstrating strong performance in automated report generation, clinical question answering, and medical image interpretation[1–5]. Recent vision-language models have shown the ability to generate radiology reports from chest radiographs, while LLMs such as GPT-5[6] and Med-PaLM 2 have achieved strong performance on medical question-answering benchmarks and clinical reasoning tasks[1–3]. In parallel, multimodal medical models such as MedGemma have extended these capabilities to image-text reasoning, supporting applications in medical visual question answering, chest X-ray interpretation, and other medical imaging tasks[4]. Beyond general benchmark performance, recent work has applied LLMs to structured radiology workflows, including information extraction from radiology reports, identification of critical findings, structured report generation, and extraction or generation of follow-up imaging recommendations[7–11]. These developments position LLMs as promising tools for augmenting radiologist decision-making and enabling scalable clinical AI workflows; however, their clinical deployment still requires careful validation, domain-specific adaptation, and safeguards to ensure reliability, transparency, and patient safety[11,12].

Despite these advances, most deployed systems still depend on single-prompting strategies, in which a single LLM instance is tasked with simultaneously performing extraction, reasoning, validation, and action generation[13]. This consolidation underutilizes the model's potential and limits interpretability[13]. In radiology, optimal decisions require context from multiple information sources, and it may take multiple steps or rounds of reasoning to form a decision[14,15].

Agentic AI systems offer a potential solution by extending LLMs beyond single-step generation toward iterative workflows in which models can plan, reason across intermediate outputs, use external tools or information sources, and adapt subsequent actions based on prior results[16]. Such systems are particularly relevant to clinical tasks that require sequential reasoning, validation, and integration of heterogeneous information rather than a single model response. However, effective agentic workflows require deliberate orchestration to define how tasks are decomposed, how information is passed between components, and how intermediate outputs are evaluated before downstream actions are taken. Multi-agent frameworks address these limitations by introducing modularity, specialization, and traceable reasoning, where agents can adaptively handle multi-step tasks by consulting multiple sources of information and building on prior agents' outputs[13–17].

In this work, we present Multi-Agent Reasoning and Coordination (MARC), a configurable multi-agent framework for clinical AI that enables structured pipeline construction without code modification, supports flexible model deployment, and introduces an automated Decomposer module for task-adaptive pipeline generation. To illustrate its flexibility, we demonstrate MARC on three representative tasks, described in detail in Section 3: biomedical question answering, radiology report generation, and task-adaptive pipeline construction.

## 2. Framework

The MARC framework is a domain-agnostic, multi-agent system implemented in Python, which leverages the LangChain library to interface with various large language models. The architecture emphasizes a sequential, modular pipeline in which specialized agents process information through explicit context passing and optional retrieval-augmented generation.

### 2.1 Multi-Agent Design

MARC employs a structured workflow design pattern (Level 2 autonomy)[20], where specialized agents execute tasks in a predetermined sequence with explicit handoffs between stages. Unlike fixed-pipeline approaches such as RadFabric[21], MARC generalizes this pattern through YAML-based configuration files that define agent sequences, model assignments, and knowledge augmentation without requiring code modifications.

Each agent encapsulates a language model, a task-specific prompt, and an optional retrieval-augmented generation (RAG) module for domain knowledge[22]. Agents are defined declaratively in a YAML configuration file rather than in code, so users can change an agent's model, prompt, or knowledge sources without modifying the underlying implementation. This declarative approach directly addresses the limitation of LLMs operating in isolation within fixed, multi-step workflows, as identified in prior single-agent medical AI systems, by enabling non-developers to modify agent workflows, swap foundation models for cost-performance optimization, and inject domain expertise through knowledge files without Python programming[13]. Full configuration details are provided in the Supplement.

### 2.2 Decomposer Module

To address the limitations of fixed pipelines and manual prompt engineering, we introduce the MARC Decomposer. Given a plain-language task description, the Decomposer employs MedGemma 4B to interpret the task, decompose it into three subtasks, and automatically generate role-specific prompt templates for each agent (Figure 1). The Decomposer operates in a single conversational turn, outputting a structured JSON specification containing agent names, roles, and full prompt templates.

Generated prompts are validated against a set of structural constraints, enforcing correct variable bindings ({input}, {previous_agent_output}), VERDICT formatting conventions, and output length restrictions, before being written to disk and loaded into the MARC pipeline. This enables dynamic pipeline construction across heterogeneous clinical tasks without manual prompt design, and generalizes MARC beyond its default radiology configuration.

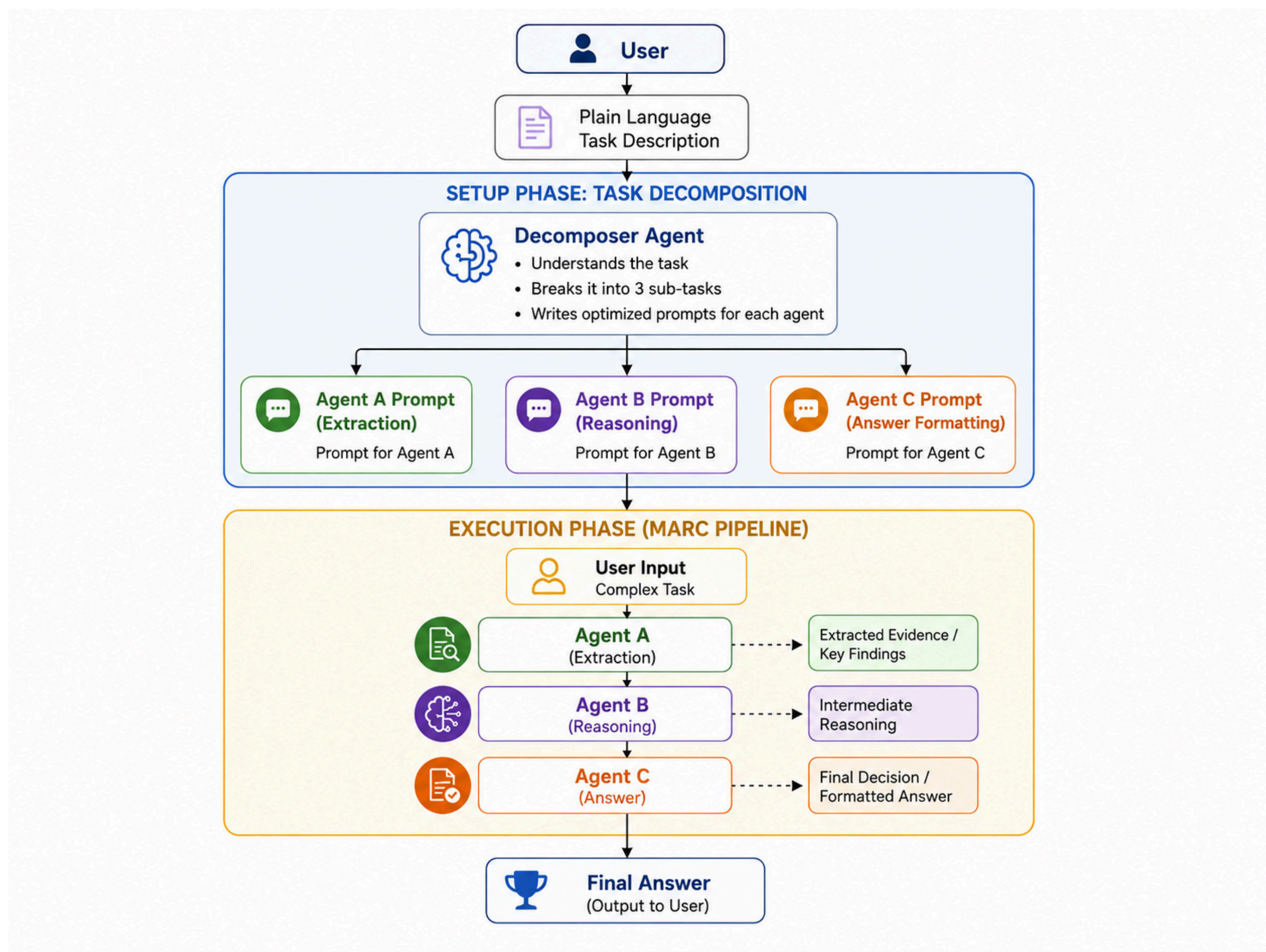

**Figure 1. MARC Decomposer workflow.** Setup Phase: the Decomposer interprets a plain-language task description and generates role-specific prompts for three agents. Execution Phase: the MARC pipeline runs with the generated prompts, passing outputs sequentially from extraction through reasoning to final answer formatting

### 2.3 Agent Prompting Strategy

All agents operate zero-shot at temperature = 0, ensuring fully deterministic and reproducible outputs across runs. Each prompt template is structured around four components: (1) a role definition establishing the agent's clinical identity and scope, (2) explicit task instructions specifying what to extract, analyze, or produce, (3) output constraints defining format, length, and valid response labels, and (4) runtime variables injected at inference time.

(A) Sequential three-agent QA pipeline
Question
Extractor Agent
Reasoner Agent
Answer Agent
Answer
context passed sequentially
(B) Radiology workflow with conditional routing
Masked Data
Unmodified Data
Tagger Agent
Healthy?
Classifier Agent
Follow-Up Agent
Evaluator Agent
User Receives Response
Input data
Agent
Decision
Output

**Figure 2. MARC pipeline architectures:** (A) Three-agent biomedical QA pipeline: Extractor → Reasoner → Answer Agent with sequential context passing. (B) Extended radiology workflow with conditional routing: abnormal reports proceed through Classifier and Follow-Up agents; all reports converge at the Evaluator.

Each agent passes its structured output to the next, so information flows cleanly from one pipeline stage to the next without agents inventing or losing prior context. The reasoning agent ends with a standardized verdict line that the final agent locates and returns verbatim, making the final prediction easy to extract without additional parsing. Full details of the variable bindings and verdict formatting are provided in the Supplement (S2).

### 2.4 Agent Specialization

MARC implements heterogeneous agent specialization through role-based prompting rather than model architecture differentiation. In the default pipeline, three specialized agents operate in sequence (Figure 2):

- **Agent 1 (Information Extraction):** Receives the raw clinical input and extracts 2-4 concise bullet points of task-relevant evidence, including key entities, clinical findings, constraints, and answer choices where applicable. The agent is explicitly instructed not to answer the question or produce a conclusion, ensuring clean separation between evidence gathering and reasoning.

- **Agent 2 (Categorization and Analysis):** Receives both the original input and Agent 1's extracted evidence. Performs structured multi-step reasoning, evaluating options or hypotheses against the extracted evidence, handling ambiguity explicitly, and producing intermediate reasoning. Terminates with a standardized verdict line encoding the final decision label.

- **Agent 3 (Recommendation Generation):** Receives Agent 2's full reasoning output and is instructed to locate the verdict line and return only its value, with no explanation or reformatting. Acts purely as a structured extractor to produce a clean, parseable final answer.

### 2.5 Agent Collaboration

The MARC pipeline executes agents in strict sequential order as defined in the configuration file. For a given input, execution proceeds as follows:

- Agent 1 receives the original user input with no prior context.
- Agent 2 receives both the original input and Agent 1's complete output.
- Agent 3 receives the original input and Agent 2's output, with implicit access to Agent 1's analysis embedded in Agent 2's response.

This chaining strategy ensures each agent operates with maximal relevant context while maintaining clear role boundaries (Figure 2). Intermediate outputs are logged at each stage, enabling post-hoc analysis of where in the pipeline errors originate.

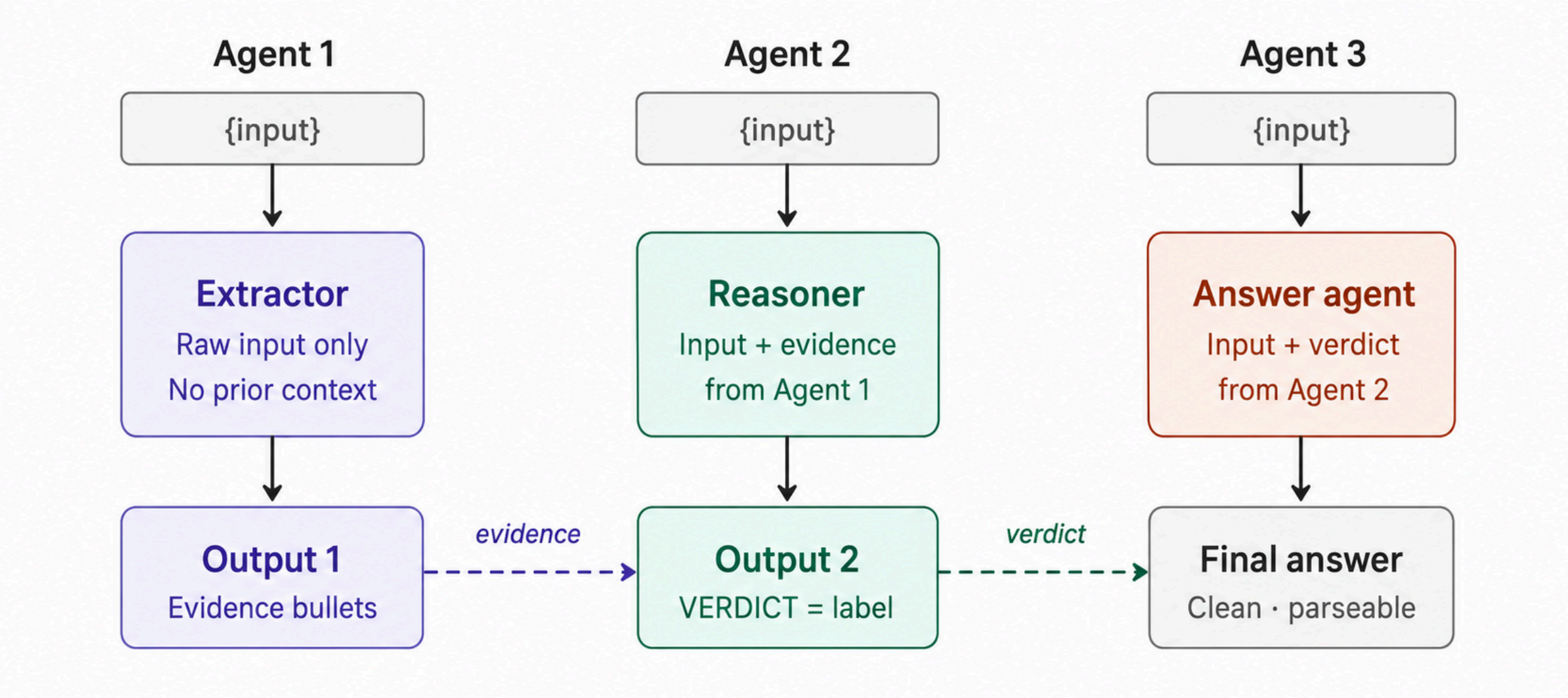

**Figure 3. Agent collaboration and context passing in the three-agent MARC pipeline:** The Extractor receives only the raw input (no prior context) and produces a set of evidence bullets (Output 1). The Reasoner receives the original input plus Agent 1's evidence and produces a verdict label (Output 2). The Answer Agent receives the original input plus Agent 2's verdict and returns a clean, parseable final answer. Intermediate outputs are logged at each stage, enabling post-hoc attribution of pipeline errors to a specific agent.

### 2.6 Deployment

MARC supports two deployment modes (Figure 4). API-based deployment uses Google's Gemini model family (gemini-2.0-flash, gemini-1.5-flash) via Google AI Studio and requires only an API key configured in the .env file. Local deployment uses Ollama for CPU-compatible inference, enabling use in institutional environments without external API dependencies. Both modes are configured identically through agents.yaml; the only difference is the model identifier specified per agent. The framework has been tested with MedGemma 4B via Ollama for clinical reasoning tasks and with Gemini models for general-purpose use.

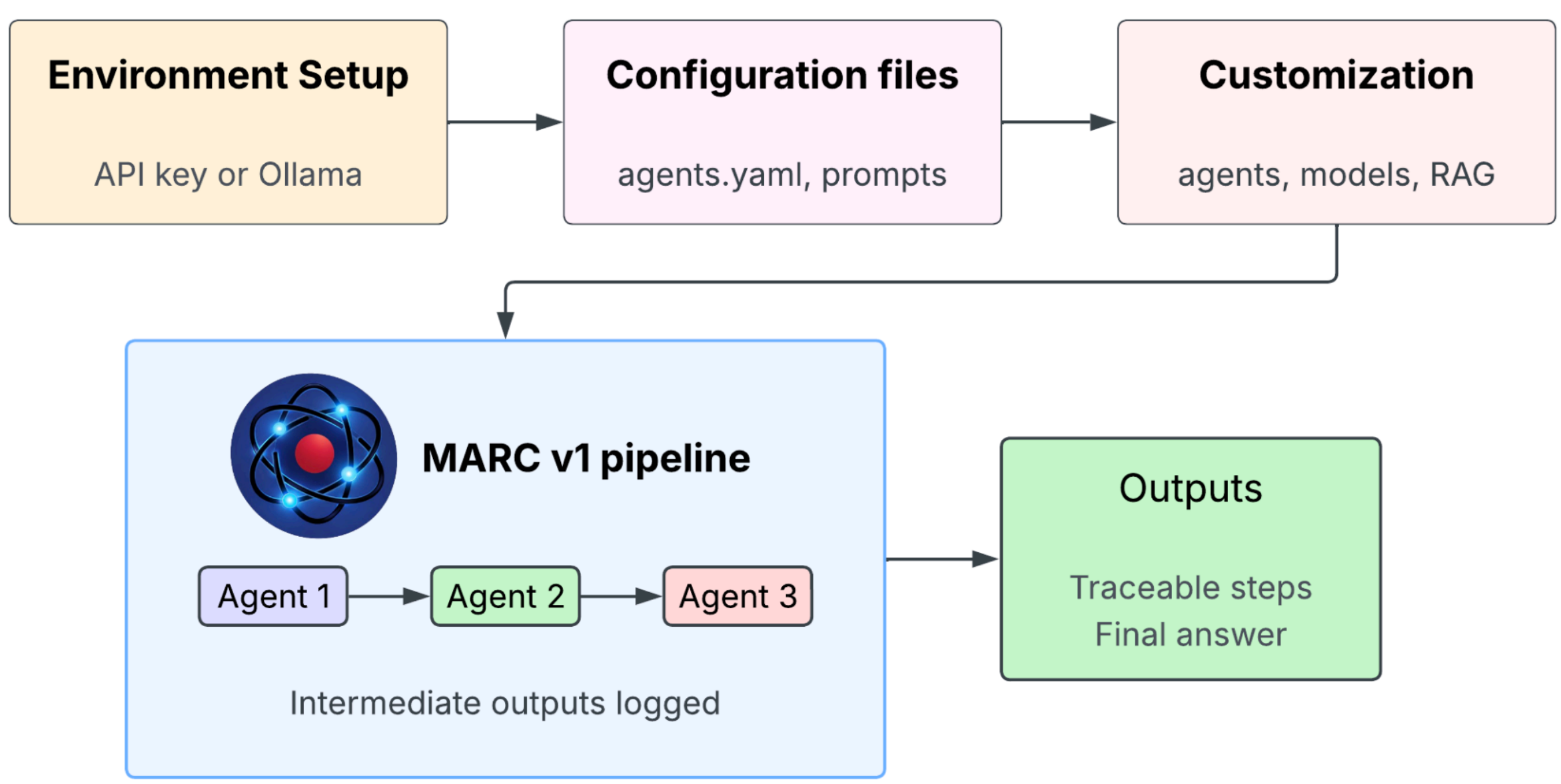

**Figure 4. MARC deployment and configuration workflow.** Users configure environment variables and model access, define agent behavior through YAML and prompt templates, customize agents, models, and optional context files, and execute the MARC pipeline to produce logged intermediate outputs and a final answer.

### 3. Use Cases

MARC is designed as a general-purpose clinical reasoning framework and can be adapted to a wide range of tasks through prompt and configuration changes alone. Three representative use cases illustrate the framework's flexibility.

- **Biomedical question answering:** In a three-agent QA pipeline, Agent 1 extracts relevant evidence from a research abstract, Agent 2 performs multi-step reasoning over the extracted evidence, and Agent 3 returns a clean yes/no/maybe verdict. This configuration maps directly to benchmarks such as PubMedQA and MedQA USMLE (Figure 2A).

- **Radiology report generation:** In an extended radiology pipeline, Agent 1 extracts clinical findings from a chest CT report, Agent 2 classifies identified pathologies, Agent 3 generates a structured impression, Agent 4 produces follow-up recommendations, and Agent 5 evaluates the full output against reference criteria (Figure 2B). Conditional routing allows normal reports to bypass classification and proceed directly to the Evaluator.

- **Task-adaptive pipeline construction via Decomposer:** For tasks outside the default radiology configuration, the Decomposer generates a complete agent pipeline from a plain-language task description. This enables rapid deployment across heterogeneous clinical tasks without manual prompt design, as demonstrated on chest CT pathology classification and USMLE-style clinical reasoning.

## 4. Discussion

Most deployed clinical LLM systems collapse several distinct cognitive steps into a single model call, which makes it difficult to determine why a system succeeds or fails. When an incorrect output is produced, the error may originate in missed information extraction, flawed reasoning, poor instruction following, or answer formatting, but these failure modes are hidden inside one opaque response. MARC instead separates these functions into role-specialized agents connected through explicit context passing, and logs each intermediate output. This enables stage-wise failure attribution and a clearer view of how a final answer was produced, which matters in clinical settings where interpretability is a practical requirement for review, debugging, validation, and trust-building.

A central contribution of MARC is that it treats orchestration as a configurable layer rather than a fixed implementation. By externalizing agent definitions, model assignments, prompt templates, and optional context files into YAML configuration files, the framework lets users modify clinical AI workflows without changing the underlying code. This lowers the barrier for clinical researchers and domain experts who understand the task and desired behavior but lack advanced programming expertise, and it supports rapid experimentation across models, prompts, and task structures. The Decomposer module extends this accessibility further by converting a plain-language task description into a structured multi-agent pipeline, reducing the need for manual prompt engineering. This is especially valuable for heterogeneous clinical tasks, where the optimal decomposition of a problem may not be obvious in advance.

MARC also addresses an important deployment concern in healthcare: institutional control over data and infrastructure. Many clinical environments face restrictions on sending patient data to external APIs, especially when working with protected health information. By supporting both API-based models and local CPU-compatible inference through Ollama, MARC can be adapted to a range of institutional settings. Local deployment with open or lightweight medical models may allow clinical teams to test and refine multi-agent workflows while maintaining greater control over data movement, cost, and reproducibility.

The framework is also model-agnostic, which is increasingly important as clinical AI moves beyond single-model evaluations. Because models can be swapped at the agent level, different stages of the pipeline can be assigned to different model families based on cost, latency, domain specialization, privacy requirements, or performance. This makes the system more adaptable as new foundation models become available.

Despite these strengths, MARC v1 has several limitations. The current default implementation emphasizes sequential agent execution, which improves interpretability but may not be optimal for all tasks. Some clinical workflows may benefit from parallel agents, dynamic routing, disagreement resolution, or iterative refinement loops. Similarly, while the three-agent architecture is useful for many reasoning tasks, more complex workflows may require variable-depth pipelines with specialized evaluators, retrievers, safety checks, or human-in-the-loop review steps. Future versions of MARC should explore more flexible orchestration strategies, including conditional branching, parallel execution, and adaptive agent selection.

Another limitation is that this manuscript primarily presents the framework design and representative use cases rather than a comprehensive empirical benchmark. Future evaluation should test MARC across diverse clinical tasks, datasets, and model backends to determine when multi-agent orchestration improves performance, reliability, interpretability, or robustness compared with single-prompt baselines. Such evaluation should include not only accuracy-based metrics but also error localization, reproducibility, latency, cost, and clinician usability. These factors are essential for determining whether multi-agent systems provide practical value in real-world clinical AI deployment.

In summary, MARC v1 provides a practical and extensible framework for building interpretable clinical AI workflows through deterministic multi-agent orchestration. Rather than treating LLMs as single-step answer generators, it structures clinical reasoning into transparent, configurable, and inspectable stages that support debugging, adaptation, local deployment, and task-specific pipeline construction. As LLMs and multimodal foundation models become more integrated into radiology and broader clinical workflows, frameworks that prioritize modularity, transparency, and institutional deployability will be essential for safe and scalable clinical AI.

## 5. Conclusion

MARC provides a practical framework for building interpretable, accessible, and institutionally deployable clinical AI systems. By separating orchestration logic from model selection, prompt design, and execution code, MARC enables clinical teams to construct, modify, and validate reasoning pipelines without requiring advanced programming expertise. Its modular multi-agent structure supports transparent intermediate outputs, flexible model deployment, and task-specific adaptation across clinical workflows. As agentic systems become increasingly integrated into radiology and broader medical practice, frameworks that prioritize transparency, configurability, and reproducibility will be essential for moving beyond monolithic prompting toward safer and more scalable clinical AI deployment.

**References:**

1. Tanno R, Barrett DGT, Sellergren A, et al. Collaboration Between Clinicians and Vision–Language Models in Radiology Report Generation. *Nat Med*. 2025;31(2):599-608. doi:10.1038/s41591-024-03302-1

2. Singhal K, Tu T, Gottweis J, et al. Toward Expert-Level Medical Question Answering with Large Language Models. *Nat Med*. 2025;31(3):943-950. doi:10.1038/s41591-024-03423-7

3. Nori H, King N, McKinney SM, Carignan D, Horvitz E. Capabilities of GPT-4 on Medical Challenge Problems. *arXiv*. Preprint posted online 2023. doi:10.48550/ARXIV.2303.13375

4. Sellergren A, Kazemzadeh S, Jaroensri T, et al. MedGemma Technical Report. *arXiv*. Preprint posted online 2025. doi:10.48550/ARXIV.2507.05201

5. Tripathi S, Sukumaran R, Cook TS. Efficient healthcare with large language models: optimizing clinical workflow and enhancing patient care. *J Am Med Inform Assoc JAMIA*. 2024;31(6):1436-1440. doi:10.1093/jamia/ocad258

6. Wang S, Hu M, Li Q, Safari M, Yang X. Capabilities of GPT-5 on Multimodal Medical Reasoning. *arXiv*. Preprint posted online August 13, 2025:arXiv:2508.08224. doi:10.48550/arXiv.2508.08224

7. Reichenpfader D, Müller H, Denecke K. A Scoping Review of Large Language Model Based Approaches for Information Extraction from Radiology Reports. *Npj Digit Med*. 2024;7(1):222. doi:10.1038/s41746-024-01219-0

8. Das A, Talati IA, Chaves JMZ, Rubin D, Banerjee I. Weakly Supervised Language Models for Automated Extraction of Critical Findings from Radiology Reports. *Npj Digit Med*. 2025;8(1):257. doi:10.1038/s41746-025-01522-4

9. Li KW, Lacson R, Guenette JP, et al. Use of ChatGPT Large Language Models to Extract Details of Recommendations for Additional Imaging From Free-Text Impressions of Radiology Reports. *Am J Roentgenol*. 2025;224(4):e2432341. doi:10.2214/AJR.24.32341

10. Wen J, Huang W, Yan H, et al. Evaluation of Large Language Models in Generating Pulmonary Nodule Follow-Up Recommendations. *Eur J Radiol Open*. 2025;14:100655. doi:10.1016/j.ejro.2025.100655

11. Busch F, Hoffmann L, Dos Santos DP, et al. Large Language Models for Structured Reporting in Radiology: Past, Present, and Future. *Eur Radiol*. 2024;35(5):2589-2602. doi:10.1007/s00330-024-11107-6

12. Bluethgen C, Van Veen D, Zakka C, et al. Best Practices for Large Language Models in Radiology. *Radiology*. 2025;315(1):e240528. doi:10.1148/radiol.240528

13. Kim TT, Makutonin M, Sirous R, Javan R. Optimizing Large Language Models in Radiology and Mitigating Pitfalls: Prompt Engineering and Fine-tuning. *RadioGraphics*. 2025;45(4):e240073. doi:10.1148/rg.240073

14. Wind S, Sopa J, Truhn D, et al. Multi-Step Retrieval and Reasoning Improves Radiology Question Answering with Large Language Models. *Npj Digit Med*. 2025;8(1):790. doi:10.1038/s41746-025-02250-5

15. Tzanis E, Adams LC, Akinci D'Antonoli T, et al. Agentic Systems in Radiology: Principles, Opportunities, Privacy Risks, Regulation, and Sustainability Concerns. *Diagn Interv Imaging*. 2026;107(1):7-16. doi:10.1016/j.diii.2025.10.002

16. Tripathi S, Cook T, Kim W. Agentic AI in Radiology. *Radiology*. 2026;318:e252730. doi:10.1148/radiol.252730

17. Salehi S, Singh Y, Horst KK, Hathaway QA, Erickson BJ. Agentic AI and Large Language Models in Radiology: Opportunities and Hallucination Challenges. *Bioengineering*. 2025;12(12):1303. doi:10.3390/bioengineering12121303

18. Faghani S, Moassefi M, Rouzrokh P, Khosravi B, Erickson BJ. Uncover This Tech Term: Agentic Artificial Intelligence in Radiology. *Korean J Radiol*. 2025;26(9):888. doi:10.3348/kjr.2025.0370

19. Khosravi B, Rouzrokh P, Akinci D'Antonoli T, et al. Agentic AI in Radiology: Evolution from Large Language Models to Future Clinical Integration. *Radiol Artif Intell*. 2026;8(2):e250651. doi:10.1148/ryai.250651

20. Schneider J. Generative to Agentic AI: Survey, Conceptualization, and Challenges. *arXiv*. Preprint posted online April 26, 2025:arXiv:2504.18875. doi:10.48550/arXiv.2504.18875

21. Chen W, Dong Y, Ding Z, et al. RadFabric: Agentic AI System with Reasoning Capability for Radiology. *arXiv*. Preprint posted online June 17, 2025:arXiv:2506.14142. doi:10.48550/arXiv.2506.14142

22. Yang R, Wong MYH, Li H, et al. Retrieval-augmented generation in medicine: A scoping review of technical implementations, clinical applications, and ethical considerations. *Cell Rep Med*. Published online July 20, 2026:102927. doi:10.1016/j.xcrm.2026.102927

## Appendix

**S1. Configuration schema.** Each agent is specified in config/agents.yaml by four parameters: agent name, model identifier, prompt template file path, and optional context file paths for RAG. Prompt templates are stored as separate text files in the prompts/ directory and loaded at runtime via YAML parsing. This separation of configuration, prompting logic, and execution code follows software engineering best practices for maintainability and reproducibility[20].

**S2. Runtime variables and verdict formatting.** Two primary variables are supported: {input}, which carries the original user query or clinical text, and {previous_agent_output}, which passes the structured output of the preceding agent. This explicit variable binding prevents agents from hallucinating prior context and enforces clean information boundaries between pipeline stages. Agent 2 prompts are additionally structured to terminate with a standardized verdict line (e.g., VERDICT = <label> or FINAL = yes/no/maybe), which Agent 3 is explicitly instructed to parse and return verbatim. This convention enforces structured handoffs and makes the final prediction extractable without post-hoc parsing heuristics.